# Effectiveness of State-of-the-Art Super Resolution Algorithms in Surveillance Environment

Muhammad Ali Farooq[1], Ammar Ali Khan[2], Ansar Ahmad[2], Rana Hammad Raza[2]

[1] National University of Ireland (NUIG), Galway, Ireland
`m.farooq3@nuigalway.ie`

[2] Pakistan Navy Engineering College (PNEC), National University of Sciences and Technology (NUST), Karachi, Pakistan
`(ammarali.khan,ansar.ahmad,hammad)@pnec.nust.edu.pk`

**Abstract.** Image Super Resolution (SR) finds applications in areas where images need to be closely inspected by the observer to extract enhanced information. One such focused application is an offline forensic analysis of surveillance feeds. Due to the limitations of camera hardware, camera pose, limited bandwidth, varying illumination conditions, and occlusions, the quality of the surveillance feed is significantly degraded at times, thereby compromising monitoring of behavior, activities, and other sporadic information in the scene. For the proposed research work, we have inspected the effectiveness of four conventional yet effective SR algorithms and three deep learning-based SR algorithms to seek the finest method that executes well in a surveillance environment with limited training data options. These algorithms generate an enhanced resolution output image from a single low-resolution (LR) input image. For performance analysis, a subset of 220 images from six surveillance datasets has been used, consisting of individuals with varying distances from the camera, changing illumination conditions, and complex backgrounds. The performance of these algorithms has been evaluated and compared using both qualitative and quantitative metrics. These SR algorithms have also been compared based on face detection accuracy. By analyzing and comparing the performance of all the algorithms, a Convolutional Neural Network (CNN) based SR technique using an external dictionary proved to be best by achieving robust face detection accuracy and scoring optimal quantitative metric results under different surveillance conditions. This is because the CNN layers progressively learn more complex features using an external dictionary.

## 1 Introduction

When low-resolution images are zoomed using upscaling techniques such as Lanczos resampling and interpolation, they become pixelated instead of providing more information. Super Resolution (SR) is an image processing technique that generates a higher resolution image from single or multiple low-resolution (LR) input images. Super

Resolution aims at adding pixel density and high-frequency content (such as textures and edges) in the LR image. SR finds tremendous utility in areas where attention to detail is of utmost importance. Some of these include medical imagery, forensic analysis of surveillance feeds, satellite images, biometric systems, and person identification, etc. This paper specifically focuses on the use of SR algorithms in offline forensic analysis of surveillance feeds. Due to its inherent nature, the surveillance environment mostly has uncontrolled and sporadic dynamics. Furthermore, the quality of a surveillance feed is affected by factors such as occlusions, type of camera hardware, camera pose, limited bandwidth, varying illumination conditions, and background complexity. These factors affect the process of identifying and monitoring individuals and various related activities in the surveillance feeds. Therefore, acquiring a generic model for varying surveillance environments is a complex task. For the proposed research work, we have inspected the effectiveness of four conventional, yet effective vision-based and three deep learning-based SR algorithms. The objective of this research work is to seek the finest method that executes well in a surveillance environment with limited training data options. To relate limited training data available for a specific surveillance environment a subset of 220 images from 6 different public datasets was selected along with some test images acquired from local settings. In an uncontrolled environment, it's complex to incorporate generic approximations that help ease further analysis. The effect of factors such as varying illumination conditions and distance from the camera concerning face detection is presented in Fig. 1. It can be observed in Fig. 1(a) that the ratio of the pixel area of the face and the whole image as a ballpark estimate is greater than 1/9 and the face is decently illuminated in an indoor scene. Contrarily, the face of the subject in Fig. 1(b) is extracted in an outdoor surveillance environment and has face shadows. Similarly, the region of interest (i.e. face) in Fig. 1(b) constitutes a very small portion of the image.

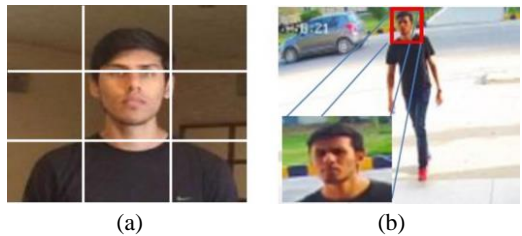

(a)          (b)

**Fig. 1.** Image acquisition environment and region of interest (a) Controlled (b) Surveillance

The rest of the paper is organized as follows. Section 2 highlights the research reported in the area of SR. The working methodology and implementation of selected SR algorithms are presented in section 3 and section 4 respectively. The qualitative and quantitative analysis and results obtained are summarized in section 5. Section 6 presents the overall conclusion drawn from the proposed work with some future research dimensions.

## 2    Background

The concept of generating a high-resolution image from an input low-resolution image is known as Super Resolution (SR). Super-resolution is not a new concept. It was first presented by Tsai [1] in 1984. Over the years, tremendous progress has been made in this area. However, super resolution remains to be an open area of research in the field of computer vision, with several new algorithms introduced every year. For a comparative analysis of super-resolution algorithms, several survey papers [2, 3, 4] have been published. These papers provide an overview of the contemporary SR techniques. To complete the discussion of SR algorithms, some of the most prominent SR techniques, both conventional and deep learning-based, are reported ahead. A selection of these will be further evaluated for their performance in a surveillance environment. Conventional SR techniques include self-similarity-based and learning-based methods among others. Yang et al. [5] proposed a learning-based SR algorithm in which relevant patches can be found when the input images are scaled-down. Kim et al. [6] have reported learning-based SR via sparse regression. Single image SR from Transformed Self-Exemplars is reported at [7] where the input image is used to generate the internal dictionary, which is then used to produce the HR image. Similarly, a self-similarity-based SR algorithm has been reported in [8]. With the advent of deep learning, several new state-of-the-art super-resolution algorithms have emerged in recent years [9, 10]. These techniques use large amounts of training data to learn the mappings between LR and HR images. Deep learning-based SR techniques include those based on Convolutional Neural Networks (CNNs) and those that employ a Generative Adversarial Network (GAN). In GAN based techniques, the model consists of a generator and a discriminator. The generator learns to produce SR images that the discriminator cannot distinguish from an actual high-resolution image. GAN based SR methods are capable of hallucinating realistic textures.

SR algorithms are generally categorized on the number of input images they require to generate the output result. A technique where the SR result is produced by just one input low-resolution image is called Single Image Single Output (SISO) Super Resolution. The practical applications of SISO involve images that require zooming of a specific area of interest e.g. subject recognition. A super-resolution technique that involves the use of multiple low-resolution input images to produce a Super Resolved result is referred to as Multi Input Single Output (MISO) Super Resolution. The applications for such techniques include satellite imagery, where multiple received LR frames are combined to find information about a particular star. Another category of super-resolution techniques involves introducing multiple low-resolution images at the input to generate multiple SR images. This is called the Multi Input Multi Output (MIMO) Super Resolution. This category mainly refers to video enhancement techniques where a stream of low-resolution video is super-resolved to give a more detailed video. An example of this category is number plate detection where the details of the number plate of a moving car are gathered via super-resolution and displayed in an enhanced video form as an output. This paper is reviewing SR algorithms with SISO.

## 3 Working Methodology of Selected SR Algorithms

In this section, we highlight and explain the working methodology of the selected algorithms. For the proposed study, we have selected seven Single Input Single Output (SISO) algorithms, of which four are conventional vision-based algorithms, and three are deep learning-based SR algorithms. The selected SR algorithms are state-of-the-art due being reported in last few years, well cited and published in top tier platforms. Among the four conventional SR algorithms, two are learning-based algorithms [5, 6], and the other two are self-similarity based algorithms [7, 8]. Whereas, regarding deep learning-based methods, two are based on a Convolutional Neural Network (CNN) [9, 10] and the remaining one leverages a Generative Adversarial Network (GAN) [11].

### 3.1 Learning-based Super Resolution

**Super Resolution via Sparse Representation (SR1).** This algorithm [5] presents a learning-based super-resolution method using sparse signal representation. Two dictionaries for LR and HR patches are trained in parallel. A sparse representation is obtained for each patch of the input low-resolution image. Coefficients of these sparse representations are used to generate a high-resolution image patch. The obtained pairs have a compact representation and are used to generate SR images. Due to local sparse modeling, the algorithm can also handle noisy images.

**Super Resolution via Sparse Regression (SR2).** It is a learning-based algorithm [6] where a patch is found from an external dictionary of LR patches and HR patches. Kernel Ridge Regression (KRR) is used in this algorithm. This algorithm uses four processing steps to give a super-resolved result. In the first step, the input low-resolution image is interpolated. In the second step, the algorithm selects the candidate images from the training data through sparse regression. Kernel Matching Pursuit (KMP) is then applied to combine the candidate image in the third step. The joined result is sharper with less noise than individual candidates. However, some artifacts are still prevalent. In the last step, post-processing takes out ringing artifacts, making the edges more pronounced. Sparse Regression converts input image patches into a sparse linear combination of elements using an external dictionary.

### 3.2 Self-Similarity-based Super Resolution

**Super Resolution from Transformed Self-Exemplars (SR3).** This method uses an internal dictionary generated from the input image itself to produce an HR image from the given LR image [7]. This algorithm relies on 'self-exemplars'. The algorithm will search examples in the down-scaled version of the input image by removing its mean. The most relevant patch is searched in the downscaled version of the input image by allowing the geometric transformation of patches. Subsequently, the match in the downscaled version is mapped in the HR patch. The patches are warped and/or rotated to find the best possible match in the downscaled version.

**Super Resolution via Patch Redundancy (SR4).** It is a self-similarity based SR algorithm [8] that identifies relevant patches from the downscaled versions of the input image, and hence an external dictionary is not required. The down-sampled image is achieved by convolving the image with a Gaussian Blur Kernel. The algorithm initiates by considering the input low-resolution patch and finding its corresponding neighbor patch using the Non-local means algorithm in the downsampled version of the image. When the neighbor patch is found, its corresponding patch in the input LR patch can be found. The corresponding patch is the super-resolved version of the input patch and is mapped in the scaled-up version of the image. The scaled-up image is achieved through Bicubic Interpolation.

### 3.3 CNN based Super Resolution

**Super Resolution using Convolutional Neural Networks (SR5).** In this method, the authors [9] propose a deep learning-based single image super-resolution method which utilizes a Convolutional Neural Network (CNN). Using an external dictionary, the algorithm learns an end-to-end mapping between LR and HR images. The proposed model takes a single LR image and converts it into an HR image through a three-layered Convolutional Neural Network (CNN). The reported model was tweaked further by the authors to investigate effect of varying number of filters towards super-resolution and related computational cost.

**Enhanced Deep Residual Networks for Single Image Super-Resolution (SR6).** This method [10] proposes residual learning-based deep super-resolution network that enhances the performance of previously proposed methods by model optimization. This involves the removal of unnecessary modules in convolutional residual networks. The authors report a single-scale super-resolution model that achieves state-of-the-art performance. They further propose a compact multiscale super-resolution model that can generate various scales of output images from a shared main network.

### 3.4 GAN based Super Resolution

**Enhanced Super-Resolution Generative Adversarial Networks (SR7).** In this paper [11], the authors propose an improved Generative Adversarial Network based SR model and is abbreviated as ESRGAN. The algorithm is a modified version of older SRGAN, aiming to achieve better perceptual quality. The authors have proposed an architecture consisting of several Residual-in-Residual Dense Blocks (RDDB) with the bottleneck layers removed to improve performance. They also introduce the use of a relativistic GAN which can learn to judge whether one image is more realistic than another image. ESRGAN achieved first place in the PIRM2018-SR challenge, an achievement that reflects its robustness.

## 4      Implementation

In this research work, a subset of 220 images from six publicly available surveillance datasets is used. Limited number of images used in this paper represent associated complexities to acquire decent image frames as training set in various surveillance environments. The dataset details are summarized below in Table 1.

Table 1. Details of six publicly available datasets used for analysis

| Dataset | | Attributes | | | |
|---|---|---|---|---|---|
| | | Frames | Subjects | Environment | Resolution |
| PETS 2006 [12] | | 30 | 5 | Airport/ Indoor | 720 x 576 |
| CAVIAR [13] | | 65 | 4 | Mall/ Indoor | 384 x 288 |
| Three Datasets from EPFL [14] | Dataset 1 | 25 | 4 | Airport/ Indoor | 480 x 270 |
| | Dataset 2 | 25 | 3 | University/ Outdoor | 360 x 288 |
| | Dataset 3 | 30 | 2 | Basement/ Indoor | 360 x 288 |
| MICC [15] | | 45 | 7 | Room/ Indoor | 426 x 320 |

The selected images consist of human subjects under different illumination and environmental conditions, time of day, and distance from the camera. The subjects are either far-field, mid-field or near field from the camera. Fig. 2 shows selected views from the mentioned datasets for quick reference of the reader.

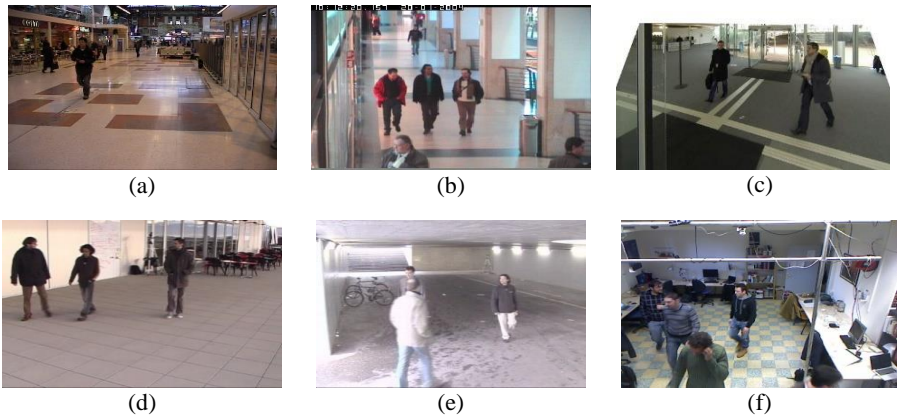

(a)           (b)           (c)

(d)           (e)           (f)

**Fig. 2.** Dataset views (a) PETS 2006 (b) CAVIAR (c) EPFL 1 (d) EPFL 2 (e) EPFL 3 (f) MICC

## 5      Experimental Results and Discussions

This section demonstrates the experimental results using the above mentioned SR algorithms. The initial performance results using these seven algorithms alongwith conventional zooming on two different locally acquired frontal face poses of male subjects are shown in Fig. 3. The input image size is 35 x 33 pixels in size.

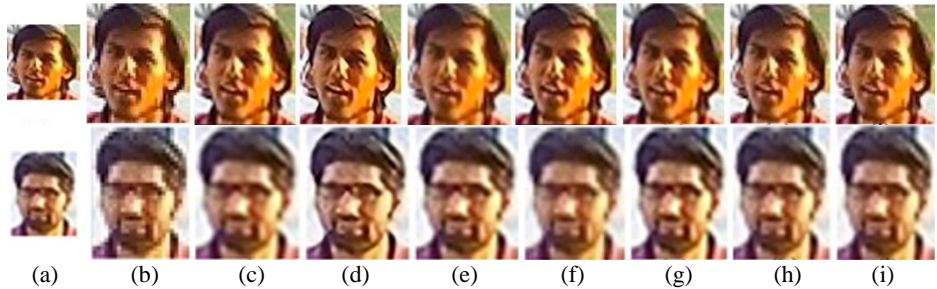

**Fig. 3.** SR results on two male subjects a) Input, b) Zoom, c) SR1, d) SR2, e) SR3, f) SR4, g) SR5, h) SR6, i) SR7

To better validate the performance of all the SR methods, all seven SR algorithms are applied to the mentioned 220 image subset to upscale the input images by a factor of four. For qualitative analysis, Multitask Cascaded Convolutional Neural Network (MTCNN) face detector [16] is applied and face detection accuracy is measured for all these datasets. For quantitative analysis, five different quantitative metrics have been utilized which include Peak Signal to Noise Ratio (PSNR), Structural Similarity Index (SSIM), Jaccard and Dice coefficients, and Face Detection Accuracy (FDA). PSNR metric is generally used to quantify the actual ratio between the maximum possible value (power) of an image and image noise that can affect the quality of image representation. SSIM is a type of perceptual metric used to measure similarity between the reference image and processed image. Jaccard and Dice coefficients are widely used for matching purposes and provide better quantitative insight into the change in structural information of the processed image. The Face Detection Accuracy (FDA) presents the MTCNN face detector's performance for the mentioned dataset images after SR processing.

To perform the analysis, all dataset frames listed in Table 1 are first down-sampled to low-resolution images followed by the application of SR methods to upscale the input image by a factor of four. The original image is then compared with the SR image. Results obtained from the quantitative analysis of the mentioned six datasets are summarized in Table 2. For the sake of simplicity, only the most and least favorable results are displayed for all datasets except CAVIAR, where all the results are shown. The best results are kept in bold.

It can be observed from Table 2 that SR5 outperforms other super-resolution methods in the majority of the datasets by achieving high face detection accuracy and robust quantitative metrics results. Whereas, SR7 Generative Adversarial Network (GAN) based technique was unable to produce exceptional results in the case of the majority of datasets which include CAVIAR, EPFL1, EPFL2, and EPFL3. Possibly this is because of poor interpretability and small sample size. The performance scale of other algorithms appears to be fluctuating in terms of good as well as poor results depending on different datasets. Overall, SR5 achieved the highest scores on qualitative and quantitative metrics on the most number of datasets. It proved to be robust under low light conditions in both, indoor and outdoor environments. Face detection using SR algorithms not only depends on the distance of the subject from the camera but face pose, environment, and illumination play a key factor as well. SR2 provided the best results under well lit and low noise conditions. This is because textures and edges are preserved when the denoising parameter is small.

**Table 2.** Quantitative results for six publicly available datasets

| Dataset | Method | PSNR (dB) | SSIM | DC % | JC % | FDA % |
|---|---|---|---|---|---|---|
| PETS2006 | SR4 | 31.656 | 0.931 | 85.1 | 89.3 | 21.8 |
| | SR5 | **33.011** | **0.986** | **90.3** | **94.9** | **50** |
| CAVIAR | Zoom | 30.799 | 0.923 | 89.4 | 92.4 | 25.5 |
| | SR1 | 32.004 | 0.949 | 95.2 | 96.8 | 48.8 |
| | SR2 | 32.276 | 0.984 | 96.1 | 97.3 | 51.1 |
| | SR3 | 32.075 | 0.976 | 95.0 | 96.6 | 45 |
| | SR4 | 31.991 | 0.938 | 92.8 | 95.3 | 36.1 |
| | SR5 | **32.491** | **0.990** | **97.8** | **98.0** | **60.5** |
| | SR6 | 20.553 | 0.840 | 94.6 | 89.8 | 16.6 |
| | SR7 | 17.737 | 0.498 | 90.8 | 83.1 | 14.6 |
| EPFL1 | SR2 | **31.747** | **0.988** | **96.4** | **97.3** | **62** |
| | SR4 | 30.671 | 0.945 | 91.2 | 94.4 | 40 |
| EPFL2 | SR4 | 31.981 | 0.911 | 89.7 | 91.3 | 37.3 |
| | SR5 | **32.981** | **0.988** | **98.8** | **99.6** | **88** |
| EPFL3 | SR4 | 30.998 | 0.899 | 91.7 | 93.4 | 33.3 |
| | SR5 | **32.291** | **0.937** | **96.5** | **98.9** | **81.6** |
| MICC | SR2 | **31.953** | **0.971** | **94.2** | **98.7** | **73.6** |
| | SR4 | 30.884 | 0.942 | 91.0 | 93.7 | 58.1 |

In some scenarios, external dictionaries can become a bit redundant if the training data does not correspond well with the input image. At times, the internal dictionary also tends to lose information when it is confronted with high-frequency elements of the image (e.g. texture). Therefore, using external or internal dictionaries have their benefits and tradeoffs.

As far as the execution of the algorithms in terms of time is concerned, example-based SR produces faster results compared to self-similarity based SR algorithms. Creating an external dictionary may take some time initially, but it generates a faster result during execution. This is mainly because self-similarity based SR algorithms have to bicubic interpolate the input LR image and then calculate the weight through non-local means for each patch in the downscaled version of the image to seek its corresponding neighbor patch. This process is relatively slower than just searching for the high-resolution patch present in the external dictionary. The execution time of the algorithm depends on the patch size chosen by the algorithm and the resolution of the input image. Higher the resolution, higher will be the computational cost and hence the time frame for execution. Smaller patch size would equate to more accuracy, as it becomes easier to search for its corresponding patches. However, the time taken for all the patches to be super-resolved increases hence the time taken for execution also increases.

For this paper, pre-trained weights of model-based SR methods have been used for analysis. The performance of these models can be further improved by training them on large surveillance datasets, thereby improving their performance on surveillance images.

## 6   Conclusion

In this study, we have analyzed the performance of seven state-of-the-art SR algorithms (i.e. 04 conventional and 03 deep learning-based). The performance of these algorithms was evaluated on a set of images gathered from six public surveillance datasets, having significant variations in lighting conditions and environments. To establish the effectiveness of each algorithm, we have used MTCNN face detector for quality metrics, as well as employed five different quantity metrics. CNN-based SR5 model even with a limited training dataset proved to be the best algorithm as it achieved robust qualitative and quantitative metrics results as compared to other algorithms. This is because the CNN layers progressively learn more complex features using an external dictionary. The results further revealed that face detection of the subjects is not only dependent upon the distance and facial pose of the subjects from the camera, but the environment and illumination conditions play a key role as well, as they affect noise levels in the image. The performance of these models can be further improved by training them on large surveillance datasets. For future work, the latest SR algorithms can be used to validate their effectiveness on thermal imaging modalities such as Long Wave Infrared (LWIR) and Near Infrared (NIR).